\documentclass{article}
\usepackage{spconf,amsmath,amssymb,bm,graphicx,booktabs,url}
\usepackage{color}
\usepackage{tikz}
\usepackage{numprint}
\usepackage{relsize}

\newcommand{\reals}{\mathbb{R}}
\newcommand{\Va}{{\mathbf{Q}}}
\newcommand{\Ha}{{\mathbf{H}}}
\newcommand{\Hak}{{\mathbf{H}_{k}}}

\newcommand{\Wa}{{\mathbf{W}}}
\newcommand{\Wak}{{\mathbf{W}_{k}}}

\newcommand{\etal}{\textit{et al. }}

\newcommand{\ie}{\textit{i.e. }}

\title{Identify, Locate and Separate: Audio-Visual Object extraction in large video Collections using weak supervision}
\name{Sanjeel Parekh$^{\star \dagger}$ ~ Alexey Ozerov$^{\dagger}$ ~ Slim Essid$^{\star}$  ~ Ngoc Q. K. Duong$^{\dagger}$ ~   Patrick P{\'e}rez$^{\mathsection}$ ~  Ga\"{e}l Richard$^{\star}$}
\address{
	$^{\star}$ LTCI, T{\'e}l{\'e}com ParisTech, Universit{\'e} Paris--Saclay, Paris, France \\
	$^{\dagger}$ Technicolor, Cesson S{\'e}vign{\'e}, France \quad
	$^{\mathsection}$ Valeo.ai, Paris, France}
\begin{document}
\ninept
\maketitle
\begin{abstract}
We tackle the problem of audio-visual scene analysis for weakly-labeled data. To this end, we build upon our previous audio-visual representation learning framework to perform object classification in noisy acoustic environments and integrate audio source enhancement capability. This is made possible by a novel use of non-negative matrix factorization for the audio modality. Our approach is founded on the multiple instance learning paradigm. Its effectiveness is established through experiments over a challenging dataset of music instrument performance videos. We also show encouraging visual object localization results.
\end{abstract}

\section{Introduction}
\label{intro}

Extracting information from audio-visual (AV) data about events, objects, sounds and scenes finds important applications in several areas such as video surveillance, multimedia indexing and robotics. Among other tasks, automatic analysis of AV scenes entails: (i) identifying events or objects, (ii) localizing them in space and time, and (iii) extracting the audio source of interest from the background. In our efforts to build a unified framework to deal with these challenging problems, we presented a first system tackling event identification and AV localization earlier \cite{parekh2018,parekhcvpr2018}. Continuing to build upon that study, in this paper we focus on making event/object classification robust to noisy acoustic environments and incorporating the ability to enhance or separate the object in the audio modality. 

There is a long history of works on supervised event detection \cite{mesaros2015sound,zhuang2010real,adavanne2017sound,bisot2017feature}. However, scaling supervision to large video collections and obtaining precise annotations for multiple tasks is both time consuming and error prone \cite{xu2015learning,gao2016segmentation}. Hence, in our previous work \cite{parekh2018} we resort to training with weak labels \ie global video-level object labels without any timing information. Multiple instance learning (MIL) is a well-known learning paradigm central to most studies using weak supervision \cite{dietterich1997solving}. MIL is typically applied to cases where labels are available over
bags (sets of instances) instead of individual instances. The task then amounts
to jointly selecting appropriate instances and estimating classifier parameters. For applying this to our case, let us begin by viewing a video as a labeled bag, containing a collection of image regions (also referred to as image proposals) and audio segments (also referred to as audio proposals) obtained by chunking the audio temporally. While such a formulation yields promising results using deep MIL  \cite{parekh2018,tian2018audio}, its audio proposal design  has two shortcomings with respect to our goals: it is (i) prone to erroneous classification in noisy acoustic conditions and (ii) limited to temporal localization of the audio event or object, thus does not allow for time-frequency segmentation in order to extract the audio source of interest. To address these shortcomings, we propose to generate audio proposals using non-negative matrix factorization (NMF) \cite{lee2001algorithms}. Note that the term \textit{proposal} refers to image or audio ``parts'' that may potentially contain the object of interest. For the audio modality these ``parts'' can be obtained through uniform chunking of the signal, as we did previously, or more sophisticated methods.

NMF is a popular unsupervised audio decomposition method that has been successfully utilized in various source separation systems \cite{ozerov2010multichannel} and as a front-end for audio event detection systems \cite{bisot2017overlapping, heittola2011sound}. It factorizes an audio spectrogram into two nonnegative matrices namely, so-called spectral patterns and their activations. Such a part-based decomposition is analogous to breaking up an image into constituent object regions. This motivates its use in our system. It makes it possible not only to de-noise the audio, but also to appropriately combine the parts for separation. An interesting work which has appeared recently uses NMF basis vectors with weak supervision from visual modality to perform audio source separation \cite{grauman2018}. There are three key differences with our proposed approach: (i) The authors of that proposal use the NMF basis vectors and not their activations for training the system. Hence no temporal information is utilized. (ii) Unlike us, they perform a supervised dictionary construction step after training to decompose a test signal (iii) Finally, they do not consider the task of visual localization. Other recent approaches for  deep learning based vision-guided audio source separation methods utilize ground-truth source masks for training \cite{sndpix,Ephrat2018}. It is worth noting that our proposed enhancement technique is significantly different as we do not use separated ground truth sources at any stage and only rely on weak labels. This makes the problem considerably more challenging.  

\textbf{Contributions.\enspace}
Building upon our inital work \cite{parekh2018}, we show how a deep MIL framework can be flexibly used to robustly perform several audio-visual scene understanding tasks using just weak labels. In particular, in addition to temporal audio proposals as in \cite{parekh2018} we propose to use NMF components as audio proposals for improved classification and to allow source enhancement.  We demonstrate the usefulness of such an approach on a large dataset of unconstrained musical instrument performance videos.  As the data is noisy, we expect NMF decomposition to provide additional, possibly ``cleaner'' information about the source of interest. Moreover, scores assigned to each component by the MIL module to indicate their relevance for classification can be reliably used to enhance or separate multiple sources.

We begin with a discussion of various modules of the proposed approach from proposal generation to classification in Section \ref{oa}. This is followed by qualitative and quantitative experimental results on classification, audio source enhancement and visual localization tasks in Section \ref{exp}.

\begin{figure*}[t!]
\centering
\input{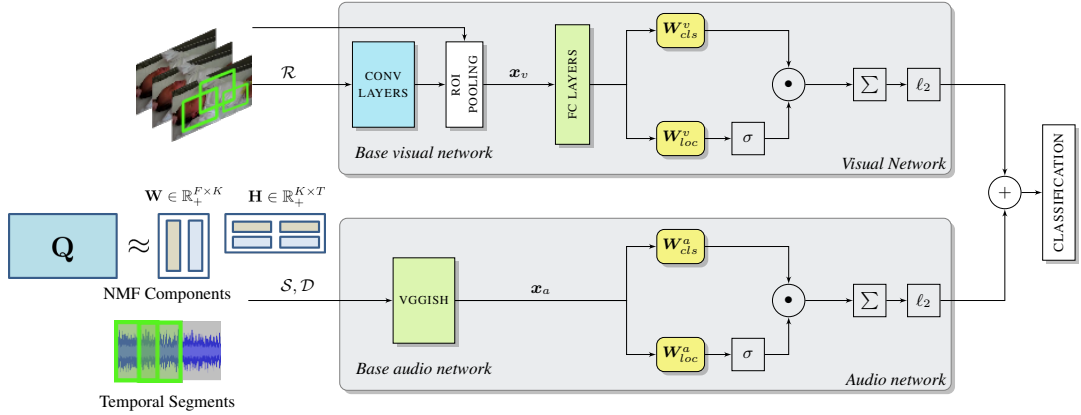}
\caption{\textbf{Proposed approach}: Given a video, we consider the depicted pipeline to go from audio and visual proposals to localization and classification. For the visual modality box proposals are considered, while for audio temporal segments and/or NMF component proposals are utilized. Weights for each module are either trained from scratch (in yellow), fine-tuned (in green) or frozen (in blue) during training.}
\label{prosys}
\end{figure*}

\section{Proposed Approach}
\label{oa}

The proposed approach is depicted in Fig. \ref{prosys}. We formulate the problem within a deep MIL framework.  Each video is considered as a bag of visual and audio proposals. These proposals are fed to their respective feature extraction and scoring networks. The scores indicate relevance of each region or segment for a particular class. Their aggregation, as depicted in Fig. \ref{prosys}, allows video-level classification. In the following section we discuss proposal generation, feature extraction, scoring and training procedures in detail.


\subsection{System Details}
\label{sd}
\textbf{Visual Proposals.\enspace} 
As our goal is to localize spatially and temporally the most discriminative image region pertaining to a class, we choose to generate proposals over video frames sub-sampled at a rate of 1 frame per second. Class-agnostic bounding-box proposals are obtained using the well-known EdgeBoxes \cite{zitnick2014edge} algorithm. To reduce the computational load and redundancy, the confidence score generated by this method is used to select top $M_{\rm img}$ proposals from each sampled image. Hence, for a 10 sec. video, such a procedure would generate a list of $M = 10 \times M_{\rm img}$ region proposals.


A fixed-length feature vector, $\bm x_v(r_m;V)  \in \reals^{d_v}$ is obtained from each image region proposal, $r_m$ in a video $V$, 
using a convolutional neural network altered with a region-of-interest (RoI) pooling layer. In practice, feature vectors $\bm x_v(\cdot)$ are passed through two fully connected layers, which are fine-tuned during training. Typically, standard CNN architectures pre-trained on ImageNet \cite{deng2009imagenet} classification are used for the purpose of initializing network weights (see Section \ref{exp} for implementation details).\\


\noindent \textbf{Audio Proposals.\enspace} 
We study two kinds of proposals: 

\begin{enumerate}
    \item \textbf{Temporal Segment Proposals (TSP)}: Herein the audio is simply decomposed into $T$ temporal segments of equal length, $\mathcal{S} = \{s_1, s_2, \ldots, s_T\}$. These proposals are obtained by transforming the raw audio waveform into log-Mel spectrogram and subsequently chunking it by sliding a fixed-length window along the temporal axis. The dimensions of this window are chosen to be compatible with the base audio network (see Sec. \ref{exp}). 
    \item \textbf{NMF Component Proposals (NCP)}: Using NMF we decompose audio magnitude spectrogram $\Va  \in \mathbb{R}_{+}^{F \times N}$ consisting of $F$ frequency bins and $N$ short-time Fourier transform (STFT) frames, such that,
		\begin{equation}
		\Va \approx \Wa \Ha,
		\end{equation}
		where $\Wa \in \mathbb{R}_{+}^{F \times K}$
		and $\Ha \in \mathbb{R}_{+}^{K \times N}$ 
		are interpreted as the nonnegative audio spectral patterns and their temporal activation matrices respectively. Here $K$ is the total number of spectral patterns. To estimate $\Wa$ and $\Ha$ we minimize the Kullback-Leibler (KL) divergence  using multiplicative update rules \cite{lee2001algorithms} where $\Wa$ and $\Ha$ are initialized randomly.

We now apply NMF-based Wiener filtering, as in \cite{fevotte2009nonnegative}, 
to an audio recording to decompose it into $K$ tracks (also referred to as NMF components) each obtained from $\Wak, \Hak$ for $k \in [1,K]$, where $\Wak$ and $\Hak$ denote spectral patterns and activations corresponding to the $k^{th}$ component, respectively. They  
can now be considered as proposals that may or may not belong to the class of interest. Specifically, we chunk each NMF component into temporal segments, which we call NMF Component proposals or NCPs. We denote the set of NCPs by $\mathcal{D} = \{d_{k,t}\}$, where each element is indexed by the component, $k \in [1,K]$ and temporal segment $t \in [1,T]$ number. As the same audio network is used for both kinds of audio proposals, for each NMF component or track we follow the TSP computation procedure. However, this is done with a non-overlapping window for reducing computational load. 
\end{enumerate}

Proposals generated by both the aforementioned methods are passed through a  VGG-style deep network known as \texttt{vggish} \cite{hershey2017cnn} for base audio feature extraction. Hershey \textit{et al}. introduced this state-of-the-art audio feature extractor as an audio counterpart to networks pre-trained on ImageNet for classification. \texttt{vggish} has been pre-trained on a preliminary version of YouTube-8M \cite{45619} 
for audio classification based on video tags. It generates a 128 dimensional embedding $\bm x_a (s_t;V) \in \reals^{128}$ for each input log-Mel spectrogram 
segment $s_t \in \reals^{96 \times 64}$ with 64 Mel-bands and 96 temporal frames. We fine-tune all the layers of \texttt{vggish} during training. 

\textbf{Proposal Scoring and Fusion. \enspace}  
Having obtained representations for each proposal in both the modalities, we now score them with respect to classes using the two-stream architecture put forth by Bilen \etal \cite{bilen2016weakly}. This module consists of parallel classification and localization streams. 
Generically denoting audio or visual proposals by $\mathcal{P}$ and their $l-$dimensional input representations to the scoring module by $Z \in \reals^{|\mathcal{P}| \times l}$, the following sequence of operations is carried out: First, $Z$ is passed through linear fully-connected layers of both classification and localization streams (shown with yellow in Fig. \ref{prosys}) giving transformed matrices $A \in \reals^{|\mathcal P| \times C}$ and $B \in \reals^{|\mathcal P| \times C}$, respectively. This is followed by a softmax operation on $B$ in the localization stream, written as: 
\begin{equation}
[\sigma(B)]_{pc} = \frac{e^{b_{pc}}}{\sum_{p=1}^{|\mathcal{P}|} e^{b_{pc}}},~\forall (p,c) \in (1,|\mathcal{P}|)\times(1,C).
\end{equation}
This allows the localization layer to choose the most relevant proposals for each class. Subsequently, the classification stream output $A$ is weighted by $\sigma(B)$ through element-wise multiplication: $E = A \odot \sigma(B)$ . Class scores over the video are obtained by summing the resulting weighted scores in $E$ over all the proposals. 

The same set of operations is carried out for both audio and visual proposals. Before addition of global level scores from both the modalities, they are $\ell_2$-normalized to ensure similar score range.

\textbf{Classification Loss and Training. \enspace} Given a set of $N$ training videos and labels, $\{(V^{(n)}, \bm y^{(n)})\}_{n=1}^{N}$, we solve a multi-label classification problem. Here $\bm y \in \mathcal{Y} = \{-1,+1\}^C$ with the class presence denoted by $+1$ and absence by $-1$. 
To recall, for each video $V^{(n)}$, the network takes as input a set of image regions $\mathcal{R}^{(n)}$ along with audio TSP $\mathcal{S}^{(n)}$, NCP $\mathcal{D}^{(n)}$ or both. After performing the described operations on each modality separately, the $\ell_2$ normalized scores are added and represented by $\phi(V^{(n)};\bm w) \in \reals^C$, with all network weights and biases denoted by $\bm w$. 
Both sub-modules are trained jointly
using the multi-label hinge loss:
\begin{equation}
L(\bm w) = \frac{1}{CN}\sum_{n=1}^N \sum_{c=1}^C \max\Big(0, 1-y^{(n)}_c\phi_c(V^{(n)};\bm w)\Big).
\end{equation}

\subsection{Source Enhancement}
As noted earlier, a by-product of training the proposed system with NCPs is the ability to perform source enhancement. This can be done by aggregating the NMF component proposal relevance scores as follows:
	\begin{itemize}
		\item Denoting by $\beta_{k,t}$ the score for $k^{th}$ component's $t^{th}$ temporal segment, we compute a global score for each component as
		\[\alpha_k = \max_{t \in T} ~ \beta_{k,t}.\]
		\item We apply min-max scaling between [0,1]:	
			\[\alpha'_k = \frac{\alpha_k - \min (\alpha)}{\max (\alpha) - \min(\alpha)}.\]
			
		\item This is followed by soft mask based source and noise spectrogram reconstruction using complex-valued mixture spectrogram $\mathbf{X}$. Note that we can optionally apply a hard threshold $\tau$ on $\alpha'_k$ to choose the top ranked components for the source. This amounts to replacing $\alpha'_k$ by the indicator function $\mathbf{1}[\alpha'_k \geq \tau]$ in the following reconstruction equations:
			\begin{equation*}
			\label{src-rec}
			    \mathbf{S} = \frac{\sum_k \alpha'_k \Wak\Hak}{\Wa\Ha} \mathbf{X}, \quad \mathbf{N} = \frac{\sum_k (1-\alpha'_k) \Wak\Hak}{\Wa\Ha} \mathbf{X}
			\end{equation*}
			
		Here $\mathbf{S}$ and $\mathbf{N}$ are the estimates of source of interest and of background noise, respectively.	
    \end{itemize}


\section{Experiments}
\label{exp}
\subsection{Setup}
\label{setup}

\textbf{Dataset.\enspace}  We use Kinetics-Instruments (KI), a subset of the Kinetics dataset \cite{kay2017kinetics} that contains 10s Youtube videos from 15 music instrument classes. From a total of 10,267 videos, we create training and testing sets that contain 9199 and 1023 videos, respectively. For source enhancement evaluation, we handpicked 45 ``clean'' instrument recordings, 3 per class. Due to their unconstrained nature, the audio recordings are mostly noisy, \ie videos are either shot with accompanying music/instruments or in acoustic environments containing other background events. In that context, ``clean'' refers to solo instrument samples with minimal amount of such noise. 


\textbf{Systems.\enspace} Based on the configuration depicted in Fig. \ref{prosys}, we propose to evaluate audio-only, A, and audio-visual (multimodal), V + A, systems with different audio proposal types, namely:
\begin{itemize}
    \item A (TSP): temporal segment proposals,
    \item A (NCP): NMF component proposals,
    \item A (TSP, NCP): all TSPs and NCPs are put together into the same bag and fed to the audio network. 
\end{itemize}
While systems using only TSP give state-of-the-art results \cite{parekh2018}, they serve as a strong baseline for establishing the usefulness of NCPs in classification. For source enhancement we compare with the following NMF related methods: 
\begin{itemize}
    \item Supervised NMF \cite{fevotte2018single}: We use the class labels to train separate dictionaries of size 100 for each music instrument with stochastic mini-batch updates. At test time, depending on the label, the mixture is projected onto the appropriate dictionary for source reconstruction.
    \item NMF Mel-Clustering \cite{spiertz2009source}: This blind audio-only method reconstructs source and noise signals by clustering mel-spectra of NMF components. We take help of the example code provided online for implementation in \textsc{matlab} \cite{melnmf}.
\end{itemize}

\textbf{Implementation Details.\enspace} All proposed systems are implemented in Tensorflow. They were trained for 10 epochs using Adam optimizer with a learning rate of $10^{-5}$ and a batch size of 1. We use the \textsc{matlab} implementation of EdgeBoxes for generating image region proposals, obtaining approximately 100 regions per video with $M_{img}=10$. 
Base visual features $\bm x_v \in \reals^{9216}$ are extracted using \texttt{caffenet} with pre-trained ImageNet weights and $6\times 6$ RoI pooling layer modification \cite{girshick2015fast}. The fully connected layers, namely $fc_6$ and $fc_7$, are fine-tuned with 50\% dropout. 

For audio, each recording is resampled to 16 kHz before processing. We use the official Tensorflow implementation of \texttt{vggish} \cite{vggish}. The whole audio network is fine-tuned during training. For TSP generation we first compute log-Mel spectrum over the whole file with a window size of 25ms and 10ms hop length. The resulting spectrum is chunked into segment proposals using a 960ms window with a 480ms stride. 
For log-Mel spectrum computation we use the accompanying \texttt{vggish} code implementation. For a 10 second recording, this yields 20 segments of size $96 \times 64$. For NCP, we consider $K=20$ components with KL divergence and multiplicative updates. As stated in Sec. \ref{sd}, each NMF component is passed through the TSP computation pipeline with a non-overlapping window, giving a total of 200 ($20 \times 10$) NCPs for a 10s audio recording.\\



\textbf{Testing Protocol} 

\begin{itemize}

\item \textit{Classification}: Kinetics-Instruments is a multi-class dataset. Hence, we consider $\text{argmax}_c s_c$ of the score vector to be the predicted class and report the overall accuracy

\item \textit{Source enhancement}: We corrupt the original audio with background noise corresponding to recordings of environments such as bus, busy street, park, etc. using one audio file per scene from the DCASE 2013 scene classification dataset \cite{stowell2015detection}. The system can be utilized in two modes: \textit{label known} and \textit{label unknown}. For the former, where the source of interest is known, we simply use the proposal ranking given by the corresponding classifier for reconstruction. For the latter, the system's classification output is used to infer the source. 

\end{itemize}

\subsection{Classification Results}
In Table \ref{cls1} we show classification results on KI for all systems explained previously. For methods using NMF decomposition, the accuracy is averaged over 5 runs to account for changes due to random initialization. We observe that the accuracies are consistent across runs \ie the standard deviation does not exceed 0.5 for any of the proposed systems.

\begin{table}[ht]
\centering
\begin{tabular}{ l l c }
\toprule
 & System & Accuracy (\%) \\ 
 \midrule
 (a) & A (TSP) & 75.3\\ 
 (b) & A (NCP) & 71.1 \\ 
 (c)& A (NCP, TSP) & 76.7 \\
 (d) &(a) + (b) & \textbf{77.3}\\
 \midrule
 (e)& V + A (TSP) & 84.5 \\
 (f)& V + A (NCP) & 80.9\\ 
 (g)& V + A (NCP, TSP) & \textbf{84.6} \\ 
 (h)& (e)  + (f) & \textbf{84.6}\\
\bottomrule
\end{tabular}
\caption{Classification results on KI test set. Here, (d) adds the classification scores of systems (a) and (b) at test time [resp. for (h)]}
\label{cls1}
\end{table}

First, we note an evident increase in performance for all the AV systems when contrasted with audio-only methods. Indeed, the image sequence provides strong complementary information about an instrument's presence when audio is noisy. We also see that using NCP in conjunction with TSP results in a noticable improvement for the audio-only systems. In comparison, this relative difference is negligible for multimodal methods. A possible explanation is that  NCPs are expected to provide complementary information in noisy acoustic conditions. Thus, their contribution in assisting TSP is visible for audio-only classification. On the other hand, vision itself serves as a strong supporting cue for classification, unaffected by noise in audio and its presence  limits the reliance on NCP. The accuracy drop when using NCP alone is expected as whole audio segments could often be easier to classify than individual components.

\begin{table}[ht]
\centering
\begin{tabular}{ c c c }
\toprule
 SNR (dB) & V + A (TSP) & V + A (NCP, TSP) \\ 
 \midrule
0 & 73.9 & \textbf{75.6}\\ 
-10 & 63.2 & \textbf{65.2} \\
-20 & 58.7 & \textbf{59.2}\\
\bottomrule
\end{tabular}
\caption{Classification accuracy on KI dataset for different levels of noise in the test audio}
\label{cls2}
\end{table}

To further test the usefulness of NCP, we corrupt the test set audio with additional noise at different SNRs using samples from DCASE 2013 scene classification data. Average classification scores over this noisy test set are reported in Table 2. We observe a clear improvement even for the multimodal system when used with NCPs. 

\subsection{Source Enhancement Results and Visual Localization Examples}

\begin{figure}[t]
    \centering
    \includegraphics[scale=0.6]{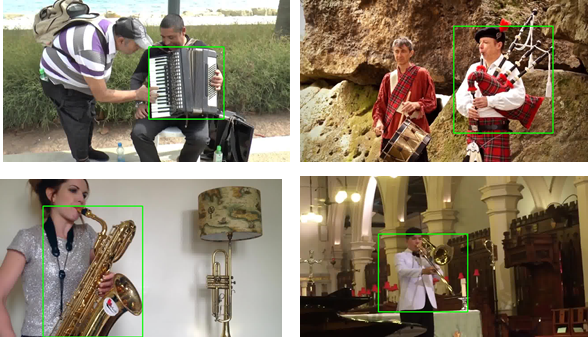}
    \caption{Visual localization examples for different instruments (clockwise from top left: accordion, bagpipes, trombone and saxophone) from the test set. Max. scoring bounding box shown in green.}
    \label{vloc}
\end{figure}

Following the testing protocol stated in Sec. \ref{setup}, we report, in Table 3, average Source to Distortion Ratio (SDR) \cite{vincent2006performance} over 450 audio mixtures created by mixing each of the 45 clean samples from the dataset with 10 noisy audio scenes. The results look promising but not state-of-the-art. This performance gap can be explained by noting that the audio network is trained for the task of audio event detection and thus does not yield optimal performance for source enhancement. The network focuses on  discriminative components, failing to separate some source components from the noise by a larger margin, possibly requiring manual thresholding for best results. Also, performance for the proposed systems does not degrade when used in ``Label Unknown'' mode, indicating that despite incorrect classification the system is able to cluster acoustically similar sounds. Performance of supervised NMF seems to suffer due to training on a noisy dataset. We present some visual localization examples in Fig. \ref{vloc}. Examples and supplementary material are available on our companion website.\footnote{\url{https://perso.telecom-paristech.fr/sparekh/icassp2019.html}} 




\begin{table}[h]
\centering
\begin{tabular}{ l c c }
\toprule
 System & Label Known & Label Unknown \\ 
 \midrule
 Supervised NMF &2.32 & -- \\
 NMF Mel-Clustering & -- & \textbf{4.32}\\
 V + A (NCP), soft & 3.29 & 3.29 \\ 
 V + A (NCP), $\tau = 0.1$ & \textbf{3.77} & 3.85 \\ 
 V + A (NCP), $\tau = 0.2$ & 3.56 & 3.56 \\
 V + A (NCP, TSP), soft & 2.11 &2.15\\
 
\bottomrule
\end{tabular}
\caption{Average SDR over mixtures created by combining clean instrument examples with environmental scenes.}
\label{sdr}
\end{table}

\section{Conclusion}
\label{con}
We have presented a novel system for robust AV object extraction under weak supervision. Unlike previous multimodal studies, we only use weak labels for training. The central idea is to perform MIL over a set of audio and visual proposals. In particular, we propose the use of NMF for generating audio proposals. Its advantage for robust AV object classification in noisy acoustic conditions and source enhancement capability is demonstrated over a large dataset of musical instrument videos.



\bibliographystyle{IEEEbib}
\bibliography{refs,av_objects,mil}

\end{document}